\documentclass{article}
\usepackage{ijcai23}

\usepackage{times}
\usepackage{soul}
\usepackage{url}
\usepackage[hidelinks]{hyperref}
\usepackage[utf8]{inputenc}
\usepackage[small]{caption}
\usepackage{graphicx}
\usepackage{amsmath}
\usepackage{amsthm}
\usepackage{booktabs}
\usepackage{algorithm}
\usepackage{algorithmic}
\usepackage[switch]{lineno}

\usepackage{subfigure}
\usepackage{multirow}
 \usepackage{amsfonts,amssymb}

\title{Boosting Decision-Based Black-Box Adversarial Attack with Gradient Priors}

\author{
Han Liu$^1$\and
Xingshuo Huang$^1$\and
Xiaotong Zhang$^1$\thanks{Corresponding author.}\and
Qimai Li$^2$\and
Fenglong Ma$^3$\and\\
Wei Wang$^4$\and
Hongyang Chen$^5$\and
Hong Yu$^1$\And
Xianchao Zhang$^{1*}$
\affiliations
$^1$Dalian University of Technology, Dalian, China\\
$^2$The Hong Kong Polytechnic University, Hong Kong, China\\
$^3$The Pennsylvania State University, Pennsylvania, USA\\
$^4$Shenzhen MSU-BIT University, Shenzhen, China\\
$^5$Zhejiang Lab, Hangzhou, China
\emails
liu.han.dut@gmail.com, \{xshuang.dut, zxt.dut\}@hotmail.com, csqmli@comp.polyu.edu.hk, fenglong@psu.edu, \{ehomewang, dr.h.chen\}@ieee.org, \{hongyu, xczhang\}@dlut.edu.cn
}

\begin{document}

\maketitle

\begin{abstract}
Decision-based methods have shown to be effective in black-box adversarial attacks, as they can obtain satisfactory performance and only require to access the final model prediction. Gradient estimation is a critical step in black-box adversarial attacks, as it will directly affect the query efficiency. Recent works have attempted to utilize gradient priors to facilitate score-based methods to obtain better results. However, these gradient priors still suffer from the edge gradient discrepancy issue and the successive iteration gradient direction issue, thus are difficult to simply extend to decision-based methods. In this paper, we propose a novel Decision-based Black-box Attack framework with Gradient Priors (DBA-GP), which seamlessly integrates the data-dependent gradient prior and time-dependent prior into the gradient estimation procedure. First, by leveraging the joint bilateral filter to deal with each random perturbation, DBA-GP can guarantee that the generated perturbations in edge locations are hardly smoothed, i.e., alleviating the edge gradient discrepancy, thus remaining the characteristics of the original image as much as possible. Second, by utilizing a new gradient updating strategy to automatically adjust the successive iteration gradient direction, DBA-GP can accelerate the convergence speed, thus improving the query efficiency. Extensive experiments have demonstrated that the proposed method outperforms other strong baselines significantly.
\end{abstract}

\section{Introduction}
Deep neural networks have achieved great success on various of tasks, such as image classification \cite{DBLP:conf/cvpr/HeZRS16,DBLP:conf/cvpr/PhamDXL21}, object detection \cite{DBLP:conf/cvpr/WangSL00021,DBLP:journals/corr/abs-2207-02696} and speech recognition \cite{DBLP:conf/icassp/ChiuSWPNCKWRGJL18,DBLP:conf/interspeech/ParkCZCZCL19}. However, recent researches demonstrate that neural networks are significantly vulnerable to adversarial examples, which are almost indistinguishable from natural data in human perception and yet classified incorrectly by the models \cite{DBLP:journals/corr/GoodfellowSS14,DBLP:conf/sp/CaoWXYFYCLL21,DBLP:conf/cvpr/ZhongLZJJ22}. This phenomenon probably causes a large risk in many real-world applications, such as spam detection \cite{DBLP:conf/acsw/WuL0X17}, automatic drive \cite{DBLP:conf/cvpr/LuoYY21,DBLP:journals/tits/MuhammadULSA21} and economic services \cite{DBLP:conf/ijcai/CintasSAOWSM20}. Investigating the generation rationale behind adversarial examples seems a promising way to improve the robustness of neural networks, which motivates the research of adversarial attacks. Based on the accessibility level of victim models, adversarial attacks can be categorized into \textbf{\emph{white-box attacks}} and \textbf{\emph{black-box attacks}}. For white-box attacks \cite{DBLP:conf/cvpr/Moosavi-Dezfooli16,DBLP:conf/sp/Carlini017}, the attackers are assumed to have full knowledge about the target model, including training data, model architecture and parameters. Therefore, it is easy to utilize gradient information to lead these methods to generate adversarial examples. However, these attack methods are overly idealistic and even impracticable in real application scenarios, as most model developers are impossible to release all the model and data information in public. For black-box attacks, the attackers only have extremely limited knowledge about the target model, e.g., the predicted labels or confidence scores, so this kind of adversarial attacks seems more promising and practical.

Existing black-box attacks mainly contain transfer-based methods, score-based methods and decision-based methods. Transfer-based methods~\cite{DBLP:conf/nips/GuoLC20,DBLP:conf/iclr/Wu0X0M20,DBLP:journals/corr/abs-2210-05968} aim to train a surrogate model to imitate the behaviors of the target model and then conduct the white-box attacks on it. This kind of attack needs a huge number of training data that are similar to the data used for training the target model, which is difficult to achieve in practice. Score-based attacks \cite{DBLP:conf/icml/GuoGYWW19,DBLP:journals/tip/LiC21,DBLP:conf/cvpr/LiJ0LZDT20} require that the target models provide the predicted scores, which is also impractical in real-world applications since they may only offer the predicted labels. Compared with transfer-based and score-based approaches, decision-based methods \cite{DBLP:conf/sp/ChenJW20,DBLP:conf/cvpr/LiXZYL20} can only use discrete predicted labels to attack the target model, thus seem more realistic and feasible. However, existing decision-based attack models are not perfect since they usually rely on a large number of queries to generate adversarial examples. 

Gradient estimation is the key point in decision-based methods, as it consumes the majority of all the queries. Recently, \cite{DBLP:conf/iclr/IlyasEM19} attempt to integrate two types of gradient priors, i.e., data-dependent prior and time-dependent prior, into score-based methods to facilitate them to obtain better performance. Nevertheless, these two priors are still difficult to simply extend to decision-based methods, as they suffer from the following drawbacks. (1) The \textbf{\emph{data-dependent prior}} follows a strong assumption, that is, if two pixels are spatially close to each other, then their estimated gradients may have similar directions. This prior only takes spatial information into consideration but ignores the importance of the values of pixels. In fact, only the pixels have similar values and are spatially close, their estimated gradients may be similar. The sharp change of pixel values usually appears on the edge of objects. Therefore, to estimate the gradients accurately, it is essential to address the edge gradient discrepancy problem. (2) The \textbf{\emph{time-dependent prior}} assumes that the gradients of successive steps are highly correlated and tend to be similar, which is suitable for score-based methods. This is because that in the iterative procedure of score-based methods, the distances between successive adversarial samples keep small, so the gradient direction of successive steps will also be similar. However, in the iterative procedure of decision-based methods, the distances between successive adversarial samples will be relatively large in the beginning, but become small subsequently. This indicates that the gradient direction of successive steps should have a similar tendency. In addition, when the similarity between estimated gradients at current and previous iterations is very large, it means that decision-based methods have fully explored in the estimated gradient direction, so a new gradient direction is needed to accelerate the convergence speed. Based on the above analysis, we need to design a crafty strategy to adjust the successive iteration gradient direction, thus boosting the query efficiency.

In this paper, we propose a novel Decision-based Black-box Attack framework with Gradient Priors (DBA-GP). To tackle the edge gradient discrepancy problem, we propose to leverage the data-dependent prior via the joint bilateral filter, which can not only smooth similar gradients for spatially close pixels with similar values, but also diversify gradients for pixels with different values. To deal with the successive iteration gradient direction problem, we simultaneously consider the distance between successive adversarial samples and the gradient direction of successive steps as additional judgment conditions, thus can generate a more appropriate gradient direction to improve the query efficiency. In summary, our contributions are as follows:
\begin{itemize}
    \item We propose a new decision-based black-box adversarial attack framework with two simple yet effective gradient priors, thus can generate high-quality adversarial examples efficiently.
    \item We discover two fundamental drawbacks of existing gradient priors, i.e., the edge gradient discrepancy issue and the successive iteration gradient direction issue. To overcome these limitations, we utilize the joint bilateral filter and two specially-designed gradient updating judgement conditions, and integrate them into decision-based attack models seamlessly. 
    \item We conduct extensive experiments against both offline and online models to validate the superiority of the proposed method compared with other strong baselines. 
\end{itemize}

\section{Related Work}
Decision-based methods only require discrete predicted labels to attack the target model, thus seem more feasible and promising in real-world applications. \cite{DBLP:conf/iclr/BrendelRB18} propose the first decision-based attack method (boundary attack), which starts with a large perturbation and then performs a random walk on the decision boundary to reduce the distance to the target image, but the use of the standard normal distribution affects the efficiency of the attack. Biased boundary attack \cite{DBLP:conf/iccv/BrunnerDTK19} uses some biases that can significantly reduce the number of queries. SIGN-OPT \cite{DBLP:conf/iclr/ChengSCC0H20} utilizes the gradient sign estimation to improve the query efficiency. EA \cite{DBLP:conf/cvpr/DongSWLL0019} designs an evolutionary algorithm to carry out the attack. HSJA \cite{DBLP:conf/sp/ChenJW20} utilizes the binary information on the decision boundary to estimate the gradient direction, which provides a fundamental and powerful framework for decision-based methods. QEBA \cite{DBLP:conf/cvpr/LiXZYL20} employs three subspace optimization methods that can reduce the number of queries and further improve the performance. PSBA \cite{DBLP:conf/icml/ZhangLLZYL21} further improves the query efficiency via progressive scaling techniques. However, it requires to train the GAN model with more additional data. SurFree \cite{DBLP:conf/cvpr/MahoFM21} attempts to move along diverse directions guided by the geometrical properties of the decision boundary. AHA \cite{DBLP:conf/iccv/LiJC0HZLLHW21} utilizes historical query information to improve the random walk optimization. Although decision-based methods have shown to be effective in adversarial attacks, they are complicated and still require a large number of queries. 

\section{Problem Formulation}
Given an input image $\textbf{\textit{x}} \in [0,1]^{dim}$, considering an $m$-class image classification model $F:\mathbb{R}^{dim} \rightarrow \mathbb{R}^m$, we can get the prediction result by $y=\text{argmax}_{i}\left[F\left(\textbf{\textit{x}}\right)\right]_i$, where $[F(\textbf{\textit{x}})]_i$ represents the probability score belonging to the $i$-th class, and $i\in\left\{ 1,2,...,m\right\}$. Given an image $\textbf{\textit{x}}^{*}$ with the true label $y^{*}$, \textbf{\emph{the targeted attack}} aims to find an adversarial image $\textbf{\textit{x}}_{adv}$ such that the model outputs a pre-specified class $y_{adv}$ under the constraint that $d\left(\textbf{\textit{x}}^*,\textbf{\textit{x}}_{adv}\right)$ is minimum, where $d(\cdot)$ is a distance measure function like $l_0$, $l_2$ or $l_{\infty}$ norm. Formally, 
\begin{equation}\label{eq1}
    \mathop{\min}\limits_{\textbf{\textit{x}}_{adv}} d\left(\textbf{\textit{x}}^*,\textbf{\textit{x}}_{adv}\right), \quad s.t., \quad \phi_{\textbf{\textit{x}}^*}(\textbf{\textit{x}}_{adv})=1,
\end{equation}
where $\phi_{\textbf{\textit{x}}^*}(\textbf{\textit{x}}):[0,1]^{dim} \longrightarrow \left\{-1,1\right\}$ is a sign function defined as:
\begin{equation}\label{eq2}
\phi_{\textbf{\textit{x}}^*}(\textbf{\textit{x}})=\text{sign}\left(S_{\textbf{\textit{x}}^*}\right)=
\begin{cases}
1& \text{if $S_{\textbf{\textit{x}}^*}(\textbf{\textit{x}})>0$,} \\
-1& \text{otherwise.} 
\end{cases}
\end{equation}
Here $S_{\textbf{\textit{x}}^*}:\mathbb{R}^{dim} \rightarrow \mathbb{R}$ is a real-valued function defined as:
\begin{equation}\label{eq3}
S_{\textbf{\textit{x}}^*}(\textbf{\textit{x}}) = [F(\textbf{\textit{x}})]_{y_{adv}}-\mathop{\max}\limits_{y \neq y_{adv}}[F(\textbf{\textit{x}})]_y.
\end{equation}

From Eq.~(\ref{eq3}), it is easy to observe that $\textbf{\textit{x}}$ is adversarial if and only if $S_{\textbf{\textit{x}}^*}(\textbf{\textit{x}}) > 0$. When $S_{\textbf{\textit{x}}^*}(\textbf{\textit{x}})=0$, $\textbf{\textit{x}}$
is exactly on the decision boundary. Note that in the decision-based black-box attack scenario, we can only get the value of function $\phi_{\textbf{\textit{x}}^*}(\textbf{\textit{x}})$. For ease of representation, hereinafter we represent the functions $S_{\textbf{\textit{x}}^*}(\textbf{\textit{x}})$ and $\phi_{\textbf{\textit{x}}^*}(\textbf{\textit{x}})$ as $S(\textbf{\textit{x}})$ and $\phi(\textbf{\textit{x}})$ respectively.

In contrast, \textbf{\emph{the untargeted attack}} aims to find an adversarial image with any incorrect category. It is worth noting that by simply treating all classes different from $y^*$ as the class $y_{others}$, we can convert an untargeted attack to a targeted attack, hence only considering the targeted attack is enough.

In terms of decision-based black-box adversarial attacks, the basic idea is to first select an initial adversarial image $\textbf{\textit{x}}_{init}$ with predicted category label $y_{adv}$, then move $\textbf{\textit{x}}_{init}$ towards $\textbf{\textit{x}}^*$ as close as possible and keep $y_{adv}$ unchanged simultaneously. In this paper, we focus on this type of method and attempt to improve this procedure with two simple yet effective gradient priors.

\section{Decision-Based Black-Box Attack with Gradient Priors}

\subsection{The Overall Framework} \label{Sec3}
The DBA-GP method utilizes a similar framework with the powerful decision-based boundary attack method HSJA \cite{DBLP:conf/sp/ChenJW20}. It adopts a sampling-based gradient estimation component to guide the search direction. Specifically, it first selects $\textbf{\textit{x}}_{init}$ as an initial adversarial image, and then performs an iterative algorithm consisting of the following three parts: (1) Estimating the gradient direction of the current adversarial image; (2) Moving the current adversarial image along the direction of the estimated gradient; (3) Approaching the decision boundary via a binary search strategy.

\subsubsection{Estimating the Gradient} 
Denote by $\textbf{\textit{x}}_{adv}^{(t)}$ the adversarial image on the decision boundary in the $t$-th iteration, the direction of the gradient $\nabla S_{\textbf{\textit{x}}^*}\left(\textbf{\textit{x}}_{adv}^{(t)}\right)$ can be approximated via the Monte Carlo estimation \cite{mooney1997monte}:
\begin{equation}\label{eq4}
\widetilde{\nabla S}\left(\textbf{\textit{x}}_{adv}^{(t)},\delta_t\right)
=\frac{1}{B}\sum_{b=1}^{B}
\phi\left(\textbf{\textit{x}}_{adv}^{(t)}+\delta_t\mathbf{u}_b\right)\mathbf{u}_b,
\end{equation}
where $\left\{\mathbf{u}_b\right\}_{b=1}^B$ are $d$-dimensional random perturbations with the unit length, and $\delta_t$ is a small positive parameter. As \cite{DBLP:conf/sp/ChenJW20} state, this estimate is accurate only if $\textbf{\textit{x}}_{adv}^{(t)}$ is on the decision boundary.

\subsubsection{Moving Along the Gradient Direction} 
When obtaining the estimated gradient, DBA-GP moves $\textbf{\textit{x}}_{adv}^{(t)}$ along the gradient direction with the following formula:
\begin{equation}\label{eq5}
\widetilde{\textbf{\textit{x}}}_{adv}^{(t)}=\textbf{\textit{x}}_{adv}^{(t)}+\xi_t\cdot
\frac{\widetilde{\nabla S}\left(\textbf{\textit{x}}_{adv}^{(t)},\delta_t\right)}
{\left\|\widetilde{\nabla S}\left(\textbf{\textit{x}}_{adv}^{(t)},\delta_t\right)\right\|_2},
\end{equation}
where $\xi_t$ is the step size at the $t$-th step, and Eq.~(\ref{eq5}) should satisfy the constraint that $\phi_{\textbf{\textit{x}}^*}\left(\widetilde{\textbf{\textit{x}}}_{adv}^{(t)}\right)=1$.

\subsubsection{Approaching the Boundary} After moving along the gradient direction, $\widetilde{\textbf{\textit{x}}}_{adv}^{(t)}$ may be far from the decision boundary. To ensure that $\textbf{\textit{x}}_{adv}^{(t+1)}$ still approaches the decision boundary, DBA-GP pulls $\widetilde{\textbf{\textit{x}}}_{adv}^{(t)}$ towards the target image $\textbf{\textit{x}}^*$ by:
\begin{equation}\label{eq6}
\textbf{\textit{x}}_{adv}^{(t+1)}=\alpha_t \cdot \textbf{\textit{x}}^*+\left(1-\alpha_t\right)
\cdot\widetilde{\textbf{\textit{x}}}_{adv}^{(t)},
\end{equation}
where the coefficient $\alpha_t \in [0,1]$, which can be determined by a binary search method.

\subsection{Gradient Estimation with Priors}
In the overall framework shown in Section \ref{Sec3},  estimating the gradient plays the most crucial role as its accuracy will directly affect the query efficiency of the method. Obviously, increasing the value of $B$ is a straightforward way to improve the quality of gradient estimation. However, in adversarial attack scenarios, as the algorithm is iterative and limited by a fixed query budget, an excessively large $B$ is unreasonable and impracticable. 

Previous studies \cite{DBLP:conf/iclr/IlyasEM19,DBLP:conf/cvpr/LiXZYL20} attempt to use data-dependent and time-dependent gradient priors to obtain more accurate gradient estimation results, which have achieved promising performance. However, they still suffer from \textbf{\emph{the edge gradient discrepancy issue}} and \textbf{\emph{the successive iteration gradient direction issue}}. In the following, we will dissect the reasons behind the above issues and propose the corresponding solutions.

\begin{figure}[t]
   \centering
    \includegraphics[width=0.46\textwidth]{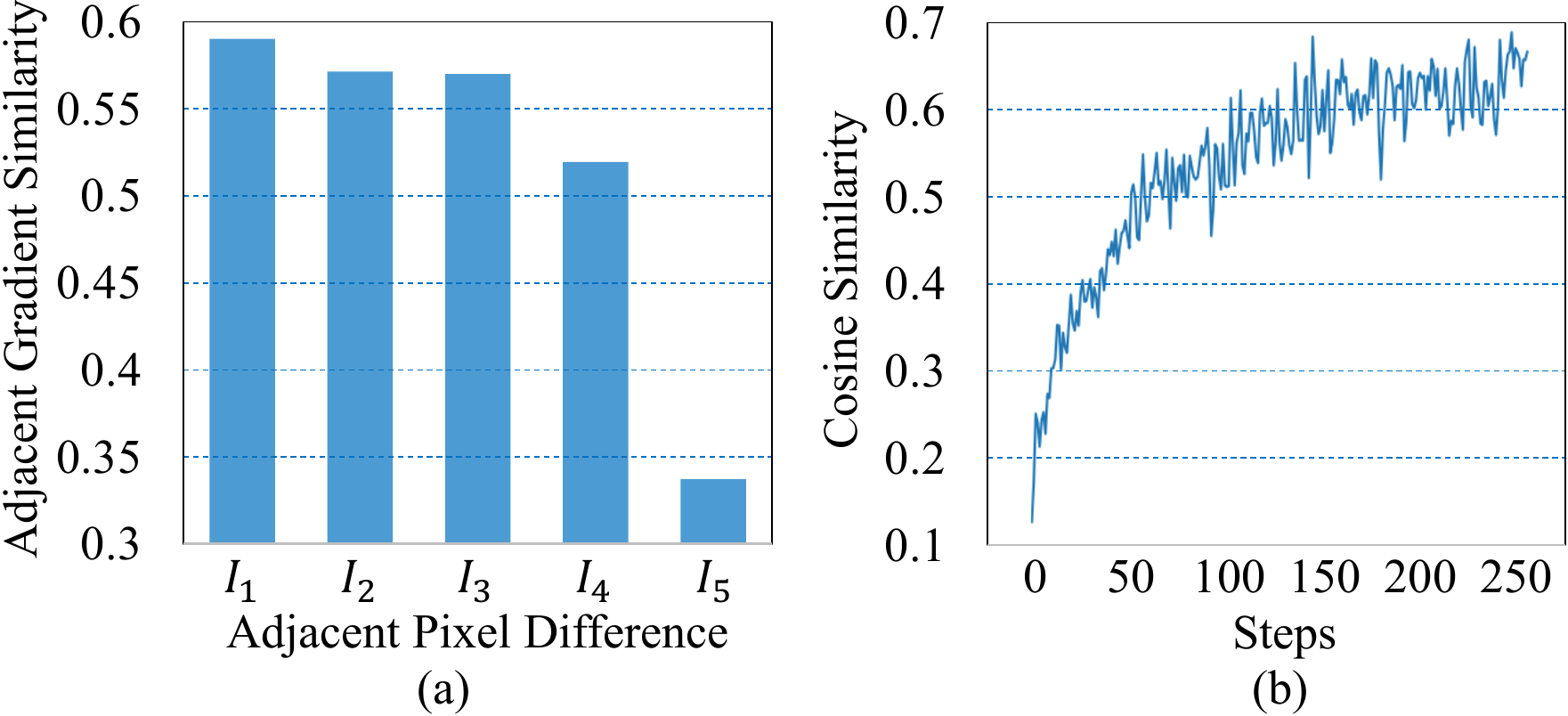}
    \captionof{figure}{(a) The ratio of similar adjacent pixel gradients in various adjacent pixel difference intervals. $I_1$, $I_2$, $\cdots$, $I_5$ represent the intervals $[0,0.2]$, $(0.2,0.4]$, $\cdots$, $(0.8,1.0]$ respectively. (b) The average cosine similarity of the gradients between the current and previous steps along the optimization trajectory of HSJA.}\label{fig:confirm}
\end{figure}

\subsubsection{Gradient Estimation with the Data-Dependent Prior}
The spatially local similarity (i.e, pixels that are close together tend to be similar) is a well-known prior in the image domain. Inspired by this fact, the data-dependent gradient prior means that the gradients of adjacent pixels tend to be similar. Specifically, if two coordinates $(i,j)$ and $(k,l)$ in $\nabla S_{\textbf{\textit{x}}^*}(\textbf{\textit{x}})$ are close, then we have $\nabla S_{\textbf{\textit{x}}^*}(\textbf{\textit{x}})_{i,j}\approx \nabla S_{\textbf{\textit{x}}^*}(\textbf{\textit{x}})_{k,l}$. 

Existing works \cite{DBLP:conf/iclr/IlyasEM19,DBLP:conf/cvpr/LiXZYL20} attempt to utilize the average-pooling, bilinear interpolation or inverse discrete cosine transform techniques to ameliorate the accuracy of gradient estimation, and have shown to be effective in black-box adversarial attacks. However, they still suffer from the edge gradient discrepancy problem. Specifically, considering the image edge locations, it is easy to discover a phenomenon that although two coordinates are close, their pixel values are usually different, then their gradients also tend to be different.

To confirm our observation, we randomly select 50 images from ImageNet \cite{DBLP:conf/cvpr/DengDSLL009}. For each image, we first normalize all pixels into $[0,1]$, calculate the absolute values of adjacent pixel differences, and assign them into five intervals $[0,0.2]$, $(0.2,0.4]$, $(0.4,0.6]$, $(0.6,0.8]$ and $(0.8,1.0]$. Then we calculate the ratio of similar adjacent pixel gradients averaged over all images. Here we treat the gradient values with the same positive and negative signs to be similar. Figure \ref{fig:confirm} (a) shows the statistic results about the ratio of similar adjacent pixel gradients in various adjacent pixel difference intervals. It can be seen that with the increase of the adjacent pixel difference, the ratio of similar adjacent pixel gradients tends to be small. That is to say, when the adjacent pixel values are close, the corresponding adjacent pixel gradients are inclined to be close. On the contrary, when the adjacent pixel values are large, the corresponding adjacent pixel gradients will be disparate. According to the above analysis, the data-dependent prior should be redefined as: the gradients of adjacent pixels tend to be similar if their pixel values are similar. 

The joint bilateral filter \cite{DBLP:journals/tog/PetschniggSACHT04} is a non-linear, edge-preserving, and noise-reducing smoothing filter, which has the following advantages. If nearby pixels are similar, it can replace the intensity of each pixel with a weighted average of intensity values from nearby pixels. If nearby pixels are diverse, it will not smooth these nearby pixels. To better leverage the data-dependent prior, we propose to use joint bilateral filter to deal with each random perturbation $\mathbf{u}_b$ with $\textbf{\textit{x}}^{*}$ as the guide image. Formally, 
\begin{equation}\label{eq7}
\widetilde{\mathbf{u}}_b=J\left(\mathbf{u}_b,\textbf{\textit{x}}^{*}\right),
\end{equation}
where $\widetilde{\mathbf{u}}_b$ is the filtered perturbation which can be used to replace $\mathbf{u}_b$. $\textbf{\textit{x}}^{*}$ is the original target image, which can provide abundant image characteristic information. $J$ is the joint bilateral filter. In the following, we give the concrete computation formula of $J$.

Given an image $A$ ($n \times n$ size) and a guided image $G$ ($n \times n$ size), after passing through the joint bilateral filter $J$, the filter output of $A_{i,j}$ can be calculated by:
\begin{equation}\label{eq8}
J(A,G)_{i,j}=\frac{1}{w(i,j)}
\sum_{(k,l)\in \Omega}
g_s(i,j,k,l)g_r(G_{i,j},G_{k,l})A_{k,l},
\end{equation}
where $\Omega$ represents a neighborhood of pixel coordinates $(i,j)$. $w(i,j)$ is a normalization term which can be obtained by:
\begin{equation}\label{eq9}
w(i,j)=\sum_{(k,l)\in \Omega}
g_s(i,j,k,l)g_r(G_{i,j},G_{k,l}).
\end{equation}
The functions $g_s(i,j,k,l)$ and $g_r(G_{i,j},G_{k,l})$ are computed by:
\begin{equation}\label{eq10}
g_s(i,j,k,l)=exp\left(-\frac{(i-k)^2+(j-l)^2}{2\sigma_s^2}\right),
\end{equation}
\begin{equation}\label{eq11}
g_r(G_{i,j},G_{k,l})=exp\left(-\frac{\left(G_{i,j}-G_{k,l}\right)^2}{2\sigma_r^2}\right),
\end{equation}
where $\sigma_s$ and $\sigma_r$ are parameters which can be used to adjust the spatial similarity and the
range (intensity/color) similarity respectively.

\subsubsection{Gradient Estimation with the Time-Dependent Prior}
The time-dependent prior means that the gradients of successive steps are heavily correlated and tend to be highly similar, which is also called the multi-step prior. \cite{DBLP:conf/iclr/IlyasEM19} attempt to extend the time-dependent prior to score-based attack methods and have achieved impressive performance. However, it is difficult to be directly applied to decision-based attack methods. The reasons are as follows. For score-based attack methods, they start from the original image and then move along the gradient direction until an adversarial image is found. Thus, the distances between successive adversarial samples in the whole iterative procedure will keep small, and the gradient direction of successive steps will also be similar. However, for decision-based attack methods, they first select an adversarial image (outside the decision boundary) which is far away from the original image, and then gradually reduce its distance from the original image. Therefore, the distances between successive adversarial samples in the iterative procedure will be relatively large in the beginning, but become small later on. In the same manner, the gradient direction of successive steps should keep a similar tendency.

To validate our hypothesis, we randomly sample 50 target images from ImageNet and calculate the average cosine similarity of the gradients between successive steps based on the decision-based boundary attack method HSJA \cite{DBLP:conf/sp/ChenJW20}. Figure~\ref{fig:confirm} (b) shows the average cosine similarity of the gradients between the current and previous steps along the optimization trajectory of HSJA. It can be seen that in the first 50 steps the gradients between successive steps are not very similar, but after that they become closer and closer.

Based on the above analysis, we attempt to use the following formula to estimate the gradient:
\begin{equation}\label{eq12}
\begin{aligned}
\widetilde{\nabla S}\left(\textbf{\textit{x}}_{adv}^{(t)},\delta_t\right)
&=\frac{1}{B}\sum_{b=1}^{B}
\phi\left(\textbf{\textit{x}}_{adv}^{(t)}+\delta_t\mathbf{u}_b\right)\mathbf{u}_b\\
&+
 \frac{1}{m}\sum_{x_{adv}^{(j)}\in\mathcal{X}^{(t)}}
\widetilde{\nabla S}\left(\textbf{\textit{x}}_{adv}^{(j)},\delta_j\right),
\end{aligned}
\end{equation}
where $\mathcal{X}^{(t)}=\{\textbf{\textit{x}}_{adv}^{(j)}:\text{max}(1,t-k)\leq j\leq t-1, D_{t,j}<\tau\}$ is the set of intermediate-process generated adversarial images that satisfy some conditions. $m$ is the number of images in $\mathcal{X}^{(t)}$. $D_{t,j}=d\left(\textbf{\textit{x}}_{adv}^{(t)},\textbf{\textit{x}}_{adv}^{(j)}\right)$ is the distance between $\textbf{\textit{x}}_{adv}^{(t)}$ to $\textbf{\textit{x}}_{adv}^{(j)}$. $k$ represents that there are $k$ iterations before the current iteration step. $\tau$ is a threshold parameter.

For Eq.~(\ref{eq12}), we first use $\mathcal{X}^{(t)}$ to filter out the intermediate-process adversarial images with large distances, and then utilize the remaining ones to facilitate the current gradient estimation. Obviously, it will increase the accuracy of the gradient estimation significantly. However, it cannot lead to an improvement in query efficiency. This is because in each iteration, decision-based attack methods will move a step along the estimated gradient direction as large as possible, which means that decision-based attack methods have fully explored in that gradient direction. Therefore, if the similarity between estimated gradients at current and previous iterations is very large, we need to generate a new gradient direction to speed up the algorithm convergence, thus improving the query efficiency. 

\begin{algorithm}[!tb]
\caption{Gradient Estimation with Priors}
\label{alg:algorithm}
\textbf{Input}: {An image $\textbf{\textit{x}}^*$, the number of random sampling $B$, the joint bilateral filter $J$, the time-dependent length $k$, the decision function $\phi$, the perturbation size $\delta_t$, the distance function $d$, the adversarial image $\textbf{\textit{x}}_{adv}^{(t)}$ in the $t$-th iteration, the adversarial image set $\{\textbf{\textit{x}}_{adv}^{(j)}: \text{max}(1,t-k)\leq j \leq t-1\}$, the gradient estimation result set $\{\widehat{\nabla S}\left(\textbf{\textit{x}}_{adv}^{(j)}, \delta_j\right):\text{max}(1,t-k)\leq j \leq t-1\}$, the threshold parameters $\tau$ and $\rho$.}\\
\textbf{Output}: {The estimated gradient $\widehat{\nabla S}\left(\textbf{\textit{x}}_{adv}^{(t)},\delta_t\right)$.}\\
\begin{algorithmic}[1]
\STATE Sample $B$ random perturbations $\{\mathbf{u}_b\}_{b=1}^B$\\
\STATE Deal with perturbations using the joint bilateral filter: 
$\{\widetilde{\mathbf{u}}_b\}_{b=1}^B=
\left\{J(\mathbf{u}_b,\textbf{\textit{x}}_{adv}^{(t)})\right\}_{b=1}^B$\\
\STATE Estimate the gradient with the following formula:
$\widetilde{\nabla S}\left(\textbf{\textit{x}}_{adv}^{(t)},\delta_t\right)=\frac{1}{B}\sum_{b=1}^B \phi\left(\textbf{\textit{x}}_{adv}^{(t)}+\delta_t\widetilde{\mathbf{u}}_b\right)\widetilde{\mathbf{u}}_b$\\
\STATE $\nabla S_{p}^{(t)}=\mathbf{0}$\\
\FOR{$j$ in $[\text{max}(1, t-k), t-1]$}
\STATE $D_{t,j}=d\left(\textbf{\textit{x}}_{adv}^{(t)},\textbf{\textit{x}}_{adv}^{(j)}\right)$ \\
\STATE $S_{t, j} = \frac{\left<\widetilde{\nabla S}\left(\textbf{\textit{x}}_{adv}^{(t)}, \delta_t\right), \widehat{\nabla S}\left(\textbf{\textit{x}}_{adv}^{(j)}, \delta_j\right)\right>}{\left\|\widetilde{\nabla S}\left(\textbf{\textit{x}}_{adv}^{(t)}, \delta_t\right)\right\|_2  \left\|\widehat{\nabla S}\left(\textbf{\textit{x}}_{adv}^{(j)}, \delta_j\right)\right\|_2}$ \\
\IF{$D_{t,j}<\tau$ and $S_{t, j}>\rho$}
\STATE {$\nabla S_{p}^{(t)} = \nabla S_{p}^{(t)} + \widehat{\nabla S}\left(\textbf{\textit{x}}_{adv}^{(j)},\delta_j\right)$}
\ENDIF
\ENDFOR
\IF {$\nabla S_{p}^{(t)}\neq\mathbf{0}$}
\STATE {
$\overline{\nabla S}_{p}^{(t)}=\frac{\nabla S_{p}^{(t)}}{\left\|\nabla S_{p}^{(t)}\right\|_2}$\\}
\ELSE
\STATE $\overline{\nabla S}_{p}^{(t)}=\mathbf{0}$\\
\ENDIF
\STATE $\widehat{\nabla S}\left(\textbf{\textit{x}}_{adv}^{(t)},\delta_t\right)=\frac{2\widetilde{\nabla S}\left(\textbf{\textit{x}}_{adv}^{(t)},\delta_t\right)}{\left\|\widetilde{\nabla S}\left(\textbf{\textit{x}}_{adv}^{(t)},\delta_t\right)\right\|}_{2} - \overline{\nabla S}_{p}^{(t)}$\\
\STATE \textbf{return} $\widehat{\nabla S}\left(\textbf{\textit{x}}_{adv}^{(t)},\delta_t\right)$
\end{algorithmic}
\end{algorithm}

To achieve the above goal, we define the set $\mathcal{A}^{(t)}=\{\textbf{\textit{x}}_{adv}^{(j)}:\text{max}(1,t-k)\leq j\leq t-1, D_{t,j}<\tau, S_{t, j} > \rho \}$, which contains intermediate-process generated adversarial images that satisfy some conditions. $D_{t,j}=d\left(\textbf{\textit{x}}_{adv}^{(t)}, \textbf{\textit{x}}_{adv}^{(j)}\right)$ is the distance between $\textbf{\textit{x}}_{adv}^{(t)}$ to $\textbf{\textit{x}}_{adv}^{(j)}$. $S_{t, j}$ is the cosine similarity between $\widetilde{\nabla S}\left(\textbf{\textit{x}}_{adv}^{(t)}, \delta_t\right)$ and $\widehat{\nabla S}\left(\textbf{\textit{x}}_{adv}^{(j)}, \delta_j\right)$, where $\widetilde{\nabla S}\left(\textbf{\textit{x}}_{adv}^{(t)}, \delta_t\right)$ is the estimated gradient in the $t$-th iteration with Eq.~(\ref{eq4}), and $\widehat{\nabla S}\left(\textbf{\textit{x}}_{adv}^{(j)}, \delta_j\right)$ is the final estimated gradient in the $j$-th iteration. $\tau$, $\rho$ and $k$ are hyperparameters. Based on the above definitions, in the $t$-th iteration we can first estimate the gradient $\widetilde{\nabla S}\left(\textbf{\textit{x}}_{adv}^{(t)}, \delta_t\right)$ with Eq.~(\ref{eq4}), and then obtain the final gradient $\widehat{\nabla S}\left(\textbf{\textit{x}}_{adv}^{(t)},\delta_t\right)$ with:
\begin{equation}\label{eq13}
\widehat{\nabla S}\left(\textbf{\textit{x}}_{adv}^{(t)},\delta_t\right)=\frac{2\widetilde{\nabla S}\left(\textbf{\textit{x}}_{adv}^{(t)},\delta_t\right)}{\left\|\widetilde{\nabla S}\left(\textbf{\textit{x}}_{adv}^{(t)},\delta_t\right)\right\|_{2}} - \overline{\nabla S}_{p}^{(t)},
\end{equation}
where $\overline{\nabla S}_{p}^{(t)}$ is the average estimated gradient of adversarial images in $\mathcal{A}^{(t)}$, and it can be calculated by:

(1) If $\mathcal{A}^{(t)}$ is not an empty set, we first compute $\nabla S_{p}^{(t)}$ by:
\begin{equation}\label{eq14}
\nabla S_{p}^{(t)}=
\sum_{x_{adv}^{(j)}\in\mathcal{A}^{(t)}}\widehat{\nabla S}\left(\textbf{\textit{x}}_{adv}^{(j)},\delta_j\right),
\end{equation}
and then compute $\overline{\nabla S}_{p}^{(t)}$ by:
\begin{equation}\label{eq15}
\overline{\nabla S}_{p}^{(t)}=\frac{\nabla S_{p}^{(t)}}{\left\|\nabla S_{p}^{(t)}\right\|_2}.
\end{equation}

(2) If $\mathcal{A}^{(t)}$ is an empty set, we simply set $\overline{\nabla S}_{p}^{(t)}=\mathbf{0}$.
By using $\widehat{\nabla S}\left(\textbf{\textit{x}}_{adv}^{(t)},\delta_t\right)$ as the estimate of the gradient direction, we can better leverage the time-dependent prior to improve the query efficiency of decision-based attack methods. 

Algorithm~\ref{alg:algorithm} summarizes the details about how to integrate data-dependent and time-dependent priors into the gradient estimation procedure.

\section{Experiments}

\subsection{Datasets}
For offline experiments, we first conduct preliminary experiments on a simple dataset MNIST~\cite{lecun1998mnist}. Then we make a comprehensive evaluation on ImageNet~\cite{DBLP:conf/cvpr/DengDSLL009} and Celeba~\cite{liu2015faceattributes} datasets. For different datasets, we exactly follow \cite{DBLP:conf/cvpr/LiXZYL20} to randomly select 50 pairs of correctly classified images from the validation set of each dataset as the target images and the initial adversarial images. For online experiments, we attack the commercial face recognition API Face++\footnote{https://www.faceplusplus.com/face-comparing/.}. 

\subsection{Victim Models}
For MNIST, we train a neural network consisting of two convolutional layers and two fully connected layers as the victim model. For ImageNet, we choose two well-known pre-trained models ResNet50~\cite{DBLP:conf/cvpr/HeZRS16} and VGG16~\cite{DBLP:journals/corr/SimonyanZ14a} as the victim models. For Celeba, we utilize samples from 100 people to fine-tune the pre-trained models ResNet50 and VGG16 on ImageNet, and take the fine-tuned models as the victim models.

\subsection{Baselines}
We compare the proposed method with the state-of-the-art targeted decision-based black-box attacks: EA~\cite{DBLP:conf/cvpr/DongSWLL0019}, 
SIGN-OPT~\cite{DBLP:conf/iclr/ChengSCC0H20}, HSJA~\cite{DBLP:conf/sp/ChenJW20}, QEBA~\cite{DBLP:conf/cvpr/LiXZYL20}, SURFREE~\cite{DBLP:conf/cvpr/MahoFM21}, and AHA~\cite{DBLP:conf/iccv/LiJC0HZLLHW21}.

\subsection{Implementation Details}

\paragraph{Evaluation metrics.}
The first evaluation metric is the average mean squared error (\textbf{MSE}) between the generated adversarial image and the target image as the number of queries increases. A smaller MSE means that the adversarial image is closer to the target image and also indicates that the attack quality is better. Under the same query budget, the lower the achieved MSE, the higher the query efficiency of the attack. The second evaluation metric is the attack success rate (\textbf{ASR}) of reaching a specified MSE threshold under a limited budget of queries. For the same query budget, a higher ASR indicates better attack quality.

\paragraph{Parameter settings.}
We develop our framework based on the FoolBox library \cite{rauber2017foolbox,DBLP:journals/jossw/RauberZBB20}. The image size of MNIST is $28 \times 28$, we set the spacial sensitivity $\sigma_s=2$ and the range sensitivity $\sigma_r=8/255$. For other datasets, we resize their image size to $3 \times 224 \times 224$, and set $\sigma_s=8$ and $\sigma_r=32/255$. We set the time-dependent length $k=5$, the MSE threshold $\tau=0.2$ and the cosine similarity threshold $\rho=0.1$ respectively. We set $B=100$, which is the number of perturbations selected in each gradient estimation. We use the $l_2$ norm as the distance measure function $d(\cdot)$. In addition, we set the step size $\xi_t=\left\|\textbf{\textit{x}}_{adv}^{(t)}-\textbf{\textit{x}}^*\right\|_2 / \sqrt{t}$ and the perturbation size $\delta_t=\left\|\textbf{\textit{x}}_{adv}^{(t)}-\textbf{\textit{x}}^*\right\|_2/dim$, where $t$ is the iteration number and $dim$ is the input dimension.

\subsection{Experimental Results}

\begin{figure}[t]
   \centering
    \includegraphics[height=4.2cm,width=5.6cm]{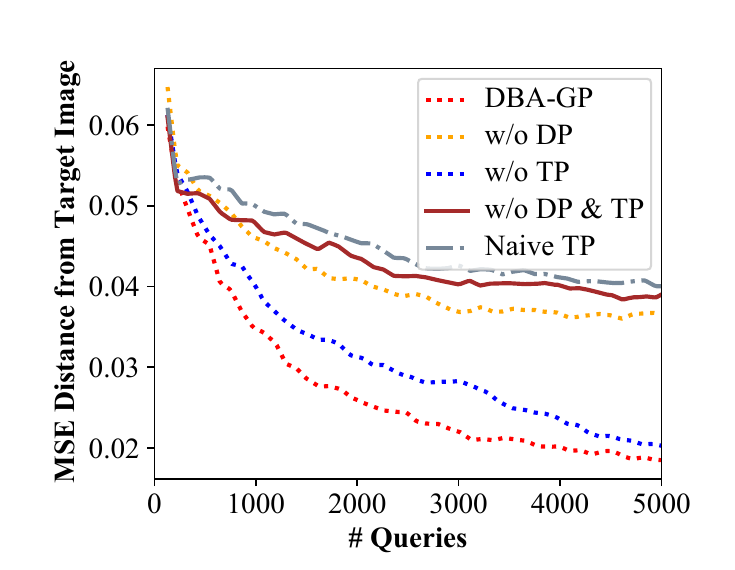}
    \captionof{figure}{The curves of MSE versus the number of queries when using different gradient priors (lower is better).}\label{fig:mnist}
     \vskip -0.1in
\end{figure}

\paragraph{Contribution analysis of each gradient prior.}
To make a comprehensive analysis of different gradient priors, we conduct the ablation study on a simple dataset MNIST, which has a smaller image resolution of $28 \times 28$, thus can converge faster and demonstrate the contribution of each gradient prior clearly. Specifically, we compare the following five cases. 
\begin{itemize}
\item \textbf{DBA-GP} means the DBA model with both time-dependent prior and data-dependent prior.
\item \textbf{Without data-dependent prior (w/o DP)} means the DBA model with only time-dependent prior. 
\item \textbf{Without time-dependent prior (w/o TP)} means the DBA model with only data-dependent prior.
\item \textbf{Without DP and TP (w/o DP \& TP)} means the DBA model without both gradient priors.
\item \textbf{With only naive time-dependent prior (Naive TP)} means the DBA model with only the naive time-dependent prior described in Eq.~(\ref{eq12}).
\end{itemize}
Figure~\ref{fig:mnist} shows the curves of MSE versus the number of queries when using different gradient priors on MNIST. $X$-axis represents the number of queries, and $Y$-axis denotes the average MSE value. From the results, we can get that w/o DP and w/o TP perform better than w/o DP \& TP, which indicates that using arbitrary gradient prior properly could improve the performance to some extent. We can also observe that the naive time-dependent prior described in Eq.~(\ref{eq12}) will bring some side effects on performance. The reason is that Eq.~(\ref{eq12}) ignores the gradient direction change at the current and previous iteration. In addition, it also can be seen that DBA-GP using both data-dependent and time-dependent gradient priors can achieve the best performance. The reason is that DBA-GP utilizes the joint bilateral filter and two specially-designed judgement conditions to better leverage data-dependent and time-dependent gradient priors respectively, thus avoiding the edge gradient discrepancy issue and the successive iteration gradient direction issue.

\begin{table*}
    \centering
    \small
    \begin{tabular}{l|c|cccccc}
        \toprule
        \multicolumn{1}{l}{Dataset} &\multicolumn{1}{l}{} &\multicolumn{3}{c}{ImageNet} & \multicolumn{3}{c}{Celeba} \\
        \cmidrule(r){1-2} \cmidrule(lr){3-5} \cmidrule(lr){6-8}
        \multicolumn{1}{l}{Method} & \multicolumn{1}{c}{Model} & Q=1K  & Q=3K  & \multicolumn{1}{c}{Q=5K}    & Q=1K  & Q=3K  & Q=5K\\
         \midrule
    
         \multirow{2}{*}{EA}        & ResNet50 & 0.0426 (0\%)  & 0.0179 (0\%)  & 0.0104 (0\%)      
                                               & 0.0693 (0\%)  & 0.0353 (0\%)  & 0.0243 (2\%)\\
                                    & VGG16    & 0.0423 (0\%)  & 0.0195 (0\%)  & 0.0123 (0\%)      
                                               & 0.0574 (0\%)  & 0.0433 (0\%)  & 0.0398 (2\%)\\
         \midrule
         \multirow{2}{*}{SIGN-OPT}  & ResNet50 & 0.0335 (0\%)  & 0.0179 (0\%)  & 0.0117 (2\%)      
                                               & 0.0409 (0\%)  & 0.0241 (2\%)  & 0.0171 (6\%)\\
                                    & VGG16    & 0.0334 (0\%)  & 0.0183 (0\%)  & 0.0119 (2\%)      
                                               & 0.0448 (0\%)  & 0.0314 (2\%)  & 0.0254 (6\%)\\
         \midrule
         \multirow{2}{*}{HSJA}      & ResNet50 & 0.0299 (0\%)  & 0.0143 (0\%)  & 0.0081 (6\%)     
                                               & 0.0377 (0\%)  & 0.0193 (6\%)  & 0.0121 (12\%)\\
                                    & VGG16    & 0.0308 (0\%)  & 0.0140 (2\%)  & 0.0080 (6\%)      
                                               & 0.0452 (4\%)  & 0.0332 (10\%)  & 0.0286 (12\%)\\
         \midrule
         \multirow{2}{*}{QEBA}      & ResNet50 & 0.0260 (0\%)  & 0.0112 (4\%)  & 0.0064 (36\%)      
                                               & 0.0124 (10\%)  & 0.0024 (48\%)  & 0.0015 (76\%)\\
                                    & VGG16    & 0.0266 (0\%)  & 0.0135 (12\%)  & 0.0096 (36\%)     
                                               & 0.0149 (18\%)  & 0.0039 (48\%)  & 0.0015 (68\%)\\
         \midrule
         \multirow{2}{*}{SURFREE}   & ResNet50 & 0.0433 (0\%)  & 0.0176 (2\%)  & 0.0116 (6\%)      
                                               & 0.0268 (0\%)  & 0.0146 (4\%)  & 0.0089 (6\%)\\
                                    & VGG16    & 0.0421 (0\%)  & 0.0233 (0\%)  & 0.0134 (6\%)      
                                               & 0.0277 (0\%)  & 0.0152 (4\%)  & 0.0081 (8\%)\\
         \midrule
         \multirow{2}{*}{AHA}       & ResNet50 & 0.0243 (0\%)  & 0.0119 (4\%)  & 0.0067 (32\%)      
                                               & 0.0119 (12\%)  & 0.0038 (42\%)  & 0.0015 (74\%)\\
                                    & VGG16    & 0.0248 (0\%)  & 0.0133 (4\%)  & 0.0099 (30\%)      
                                               & 0.0138 (18\%)  & 0.0043 (46\%)  & 0.0021 (60\%)\\
         \midrule
         \multirow{2}{*}{DBA-GP}    & ResNet50 & \textbf{0.0190 (0\%)}  & \textbf{0.0056 (20\%)}  & \textbf{0.0025 (48\%)}  
                                               & \textbf{0.0047 (30\%)}  & \textbf{0.0007 (88\%)}  & \textbf{0.0004 (92\%)}\\
                                    & VGG16    & \textbf{0.0214 (2\%)}  & \textbf{0.0075 (14\%)}  & \textbf{0.0036 (36\%)}      
                                               & \textbf{0.0049 (30\%)}  & \textbf{0.0009 (66\%)}  & \textbf{0.0005 (86\%)}\\  
        \bottomrule
    \end{tabular}
    \caption{Attack performance against different target models on different datasets. The main results in the table are the MSE values between the adversarial and target images, and the ASR values are shown in parentheses.}
    \label{tab:main}
\end{table*}

\begin{table}
\small
    \centering
    \begin{tabular}{lccccccccccccc}
        \toprule
        \multicolumn{1}{l}{Method}  & Q=1K & Q=3K & Q=5K\\
         \midrule
         EA       & 0.0926 (0\%) & 0.0910 (0\%) & 0.0893 (0\%)\\
         SIGN-OPT & 0.0833 (0\%) & 0.0801 (0\%) & 0.0764 (0\%)\\
         HSJA     & 0.0684 (0\%) & 0.0651 (0\%) & 0.0634 (0\%) \\
         QEBA     & 0.0583 (4\%) & 0.0493 (4\%) & 0.0427 (8\%) \\
         SURFREE  & 0.0799 (0\%) & 0.0741 (0\%) & 0.0689 (0\%)\\
         AHA      & 0.0621 (0\%) & 0.0542 (2\%) & 0.0468 (6\%)\\
         DBA-GP   & \textbf{0.0415 (6\%)} & \textbf{0.0272 (12\%)} & \textbf{0.0205 (22\%)} \\
      \bottomrule
    \end{tabular}
    \caption{Attack performance of defending with adversarial training on ImageNet.}
    \label{tab:def}
     \vskip -0.1in
\end{table}

\paragraph{Comparison with baselines.}
Table~\ref{tab:main} shows the performance against different baselines on ImageNet and Celeba with 1K, 3K, and 5K queries respectively. The main results in the table are MSE values, and the results in parentheses represent ASR values when the MSE threshold is 0.001. The best results are highlighted in bold. It can be observed that DBA-GP performs much better than other strong baselines. Specifically, in terms of MSE under 3K queries, no matter attacking ResNet50 or VGG16, DBA-GP can have less than one-half of MSE values than other models on ImageNet, and even less than one-third of MSE values on Celeba. In terms of ASR with 3K queries, when attacking ResNet50 on ImageNet and Celeba, DBA-GP can improve the ASR values by 16\% and 40\% respectively compared with the best baseline. The reason is that DBA-GP utilizes more gradient prior information, which enables it can obtain better adversarial images within less number of queries. In addition, as the query number increases, each method tends to achieve a lower MSE and a higher ASR. But the convergence speed of DBA-GP is the fastest, which makes it more promising in real applications.

\begin{figure}[t]
   \centering
    \includegraphics[width=7.9cm]{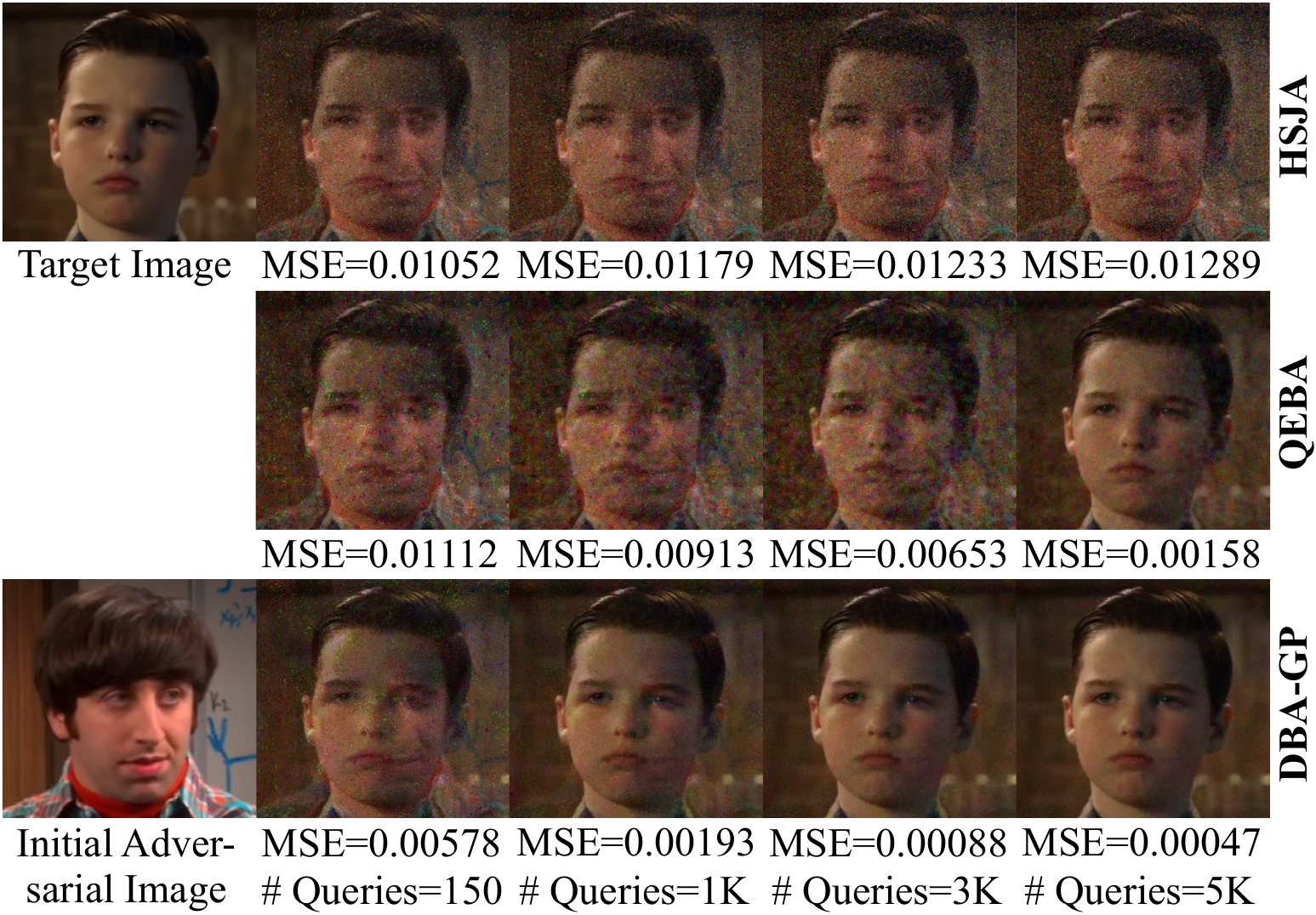}
    \captionof{figure}{Comparison of different attack models when attacking against real-world API Face++.}\label{fig:api}
    \vskip -0.1in
\end{figure}

\paragraph{Attack the real-world API Face++.}
We also attack the real-world face recognition API Face++ from MEGVII. The API  Face++ could give the prediction confidence score of whether two images contain the same person. In the experiment, when the returned confidence score is greater than 60\%, we think the corresponding images are labeled as the same person. Note that since the pixel values of the images uploaded to the API are $8$-bit floating point numbers that are not continuous in $[0, 1]$, we follow QEBA~\cite{DBLP:conf/cvpr/LiXZYL20} to discretize the images. Figure~\ref{fig:api} shows the results of attacking Face++ API. The first column is the target image and the initial adversarial image, and the last four columns are the adversarial images produced by different attack methods under different query numbers. We can observe that the adversarial image attempts to get close to the target image gradually and keep the label unchanged. For HSJA, the MSE values do not decrease with increasing the number of queries, which indicates that it is difficult to find a better adversarial image. In addition, the right side of the generated adversarial image always contains the facial feature of the initial adversarial image, which also indicates HSJA works not well. QEBA performs much better, but it requires 5K queries to get a clean-looking adversarial image. DBA-GP could generate high-quality adversarial images at only 1K queries, and the MSE value decreases continually as the number of queries increases. All these phenomena validate the superiority of our proposed DBA-GP.

\paragraph{Attack results of defending with adversarial training.}
Adversarial training~\cite{DBLP:conf/iclr/MadryMSTV18} has shown to be effective to defend against adversarial attacks. Therefore, we further compare the performance of different attacking models when attacking the ResNet-152 model with adversarial training on ImageNet. Table~\ref{tab:def} gives the results, the main results are MSE values, and the results in parentheses represent ASR values when the MSE threshold is 0.01. The results in Table~\ref{tab:def} show that our method DBA-GP can also obtain the best adversarial performance.

\section{Conclusion}
In this paper, we propose DBA-GP, a novel decision-based black-box attack framework with gradient priors. To better leverage the data-dependent gradient prior, DBA-GP exploits the joint bilateral filter to process each perturbation, which can mitigate the edge gradient discrepancy to some extent. To seamlessly integrate with the time-dependent gradient prior, DBA-GP introduces a new successive iteration gradient direction update strategy, which can speed up the convergence significantly. Extensive experiments confirm the superiority of our proposed method over other strong baselines, especially in query efficiency. In future work, we plan to investigate the theoretical underpinnings of the proposed method and extend it to other types of black-box adversarial attack methods.

\section*{Acknowledgments}
The authors are grateful to the reviewers for their valuable comments. This work was supported by National Natural Science Foundation of China (No. 62106035, 62206038, 61972065) and Fundamental Research Funds for the Central Universities (No. DUT20RC(3)040, DUT20RC(3)066), and supported in part by Key Research Project of Zhejiang Lab (No. 2022PI0AC01), National Key Research and Development Program of China (2022YFB4500300) and CAAI-Huawei Mindspore Open Fund. We also would like to thank Dalian Ascend AI Computing Center and Dalian Ascend AI Ecosystem Innovation Center for providing inclusive computing power and technical support.

\bibliographystyle{named}
\bibliography{ijcai23}

\end{document}